\documentclass[letterpaper, 10 pt, conference]{ieeeconf}

\IEEEoverridecommandlockouts
\usepackage{graphicx} 
\usepackage{subcaption} 
\usepackage{multicol} 
\usepackage{afterpage} 

\usepackage{fancyhdr}
\fancypagestyle{withfooter}{
  
  \fancyfoot[C]{\footnotesize Accepted to the IEEE ICRA Workshop on Field Robotics 2024}
}
\fancypagestyle{withfooter2}{
  
  \fancyfoot[C]{\raisebox{-339pt}{\footnotesize Accepted to the IEEE ICRA Workshop on Field Robotics 2024}}
}

\title{\LARGE \bf
Robotic deployment on construction sites: considerations for safety and productivity impact
}

\author{ Rafael Gomes Braga$^{1}$, Muhammad Owais Tahir$^{1}$, Ivanka Iordanova$^{2}$ and David St-Onge$^{1}$.
\thanks{$^{1}$ are with the Lab of INIT Robots, Mechanical Engineering, Ecole de technologie supérieure, 1100 Notre-Dame W., Canada
        {\tt\small name.surname@etsmtl.ca}}%
\thanks{$^{2}$ is with GRIDD, Construction Engineering, Ecole de technologie supérieure, 1100 Notre-Dame W., Canada
        {\tt\small name.surname@etsmtl.ca}}%
}

\begin{document}

\maketitle
\thispagestyle{withfooter}
\pagestyle{withfooter}

\begin{abstract}

Deploying mobile robots in construction sites to collaborate with workers or perform automated tasks such as surveillance and inspections carries the potential to greatly increase productivity, reduce human errors, and save costs. However ensuring human safety is a major concern, and the rough and dynamic construction environments pose multiple challenges for robot deployment. In this paper, we present the insights we obtained from our collaborations with construction companies in Canada and discuss our experiences deploying a semi-autonomous mobile robot in real construction scenarios.

\end{abstract}

\section{INTRODUCTION}

The imperative of ensuring human safety in workplaces shared with robotic systems presents a significant challenge for the field deployment of such systems. Risks associated with human-robot collaboration encompass both direct and indirect hazards. Direct risks arise from collisions between the platform and obstacles, such as equipment or personnel, while indirect risks stem from environmental factors that affect robot functionality, including noise and wind, as well as potential distractions posed by the presence of the robot itself. Addressing these concerns, Moud, Hashem Izadi, et al. proposed a model for evaluating the safety of mobile robot deployment based on physical characteristics and environmental considerations \cite{moud2020safety}.

Similarly, Augustsson et al. advocated for the establishment of defined safety zones around robots to facilitate safe interactions with humans, with the flexibility of zone geometries tailored to accommodate varying work scenarios \cite{augustsson2014human}. Complementing this approach, Truong et al. emphasized the importance of human safety in shared environments by implementing a system wherein robots dynamically generate safe navigation zones upon detecting individuals through sensor feedback \cite{truong2014dynamic}. Additionally, Shin et al. proposed a framework aligned with ISO/TS 15066 to evaluate the peak pressure of collisions involving collaborative robots in proximity to humans, underlining the critical importance of safety considerations within shared workspaces \cite{shin2019real}.

Our work focuses on the construction industry, recognized for its inherent hazards to human workers. It presents a particularly challenging environment for the deployment of autonomous mobile robots. Kim et al. highlighted persistent safety concerns surrounding human-robot collisions at construction sites, accentuated by the mobility of the robots \cite{Kim2020ProximityConstruction}. While early endeavors in human-robot collaboration within construction settings have demonstrated potential efficiency gains, the complexities of maintaining safety alongside productivity remain significant \cite{khatib1987unified}.

The objective of this paper is to study the safety and productivity impacts of deploying autonomous mobile robots on construction sites, as perceived in a real industry context. To this end, we first review the industry standards on this subject, followed by directed interviews with construction industry experts in mobile robotics. Finally, we share our experience at deploying a mobile robot prototype specifically equipped for monitoring progress at construction sites. 

\begin{figure}
    \centering
    \includegraphics[width=.5\columnwidth]{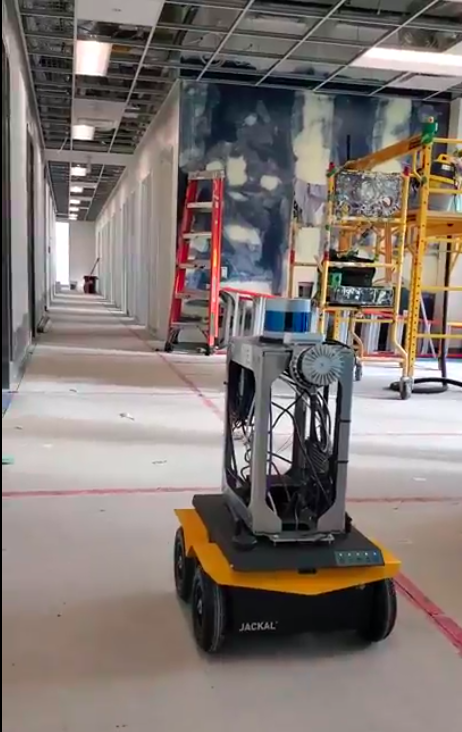}
    \caption{Our autonomous mobile robot, Journeybot, navigating a construction site}
    \label{fig:journeybot_on_site}
\end{figure}

\section{Industry Standards}
Among the most recent standards, ANSI/RIA R15.08 serves as a pivotal framework for regulating industrial mobile robots (IMRs). This standard encompasses both autonomous mobile robots (AMRs) and automated guided vehicles (AGVs) within industrial settings. While AGVs follow predetermined routes, AMRs leverage sensor technologies to navigate autonomously around obstacles. Rooted in principles from ANSI/RIA R15.06, which addresses industrial robot safety, and ANSI/ITSDF B56.5, which pertains to guided industrial vehicles, ANSI/RIA R15.08 outlines comprehensive requirements for IMRs, including operational modes, control functionalities, and presence-sensing devices.

In addition to industry standards, compliance with local regulations, such as the \emph{Safety Code for the construction industry}\footnote{https://www.legisquebec.gouv.qc.ca/en/document/cr/s-2.1,\%20r.\%204} in Quebec, Canada, is critical, particularly in construction environments where health and safety standards are paramount. Local and national codes often supersede industry standards, mandating strict adherence. On construction sites, a Safety Agent oversees compliance with these regulations to ensure worker safety.

Traffic control is essential on construction sites to prevent collisions between pedestrians and vehicles. The project manager must develop a traffic plan to restrict backing maneuvers and implement safety measures to protect individuals on the site. This plan should include provisions for any IMRs deployed. Site signalers must wear high-visibility clothing, remain visible to vehicle drivers, and stay out of their path. Consequently, workers must also be visible and identifiable to IMRs by wearing high-visibility clothing.

Self-propelled vehicles must be equipped with effective brakes and warning devices, used when approaching pedestrians, doors, turns, and hazardous locations. These requirements ensure that vehicles are operated safely, especially in the dynamic environment of a construction site. Only experienced drivers, or those under supervision, should operate such equipment. Additionally, all self-propelled vehicles must have an audible warning device within the driver’s reach, distinct from other site signals, and loud enough to be heard over construction noise.

These regulatory measures not only ensure the operational efficiency of mobile robots but also facilitate their safe integration into the dynamic and complex environments of construction sites. However, adhering to standards alone is insufficient to guarantee the safety of these robotic deployments; it is equally important to secure the acceptance of the technology by the workers.

\section{Insights from the industry}
Mobile robots have a wide range of usages for construction work, with drones (mainly outdoor ones) by far the most popular robots currently deployed. Several of their applications outdoor are envisioned for future indoor mobile robots, such as mapping, surveying, inspections, and marketing purposes. Indoor ground robots would be mainly used for progress monitoring and quality control. Other more specialized robots - equipped with IoT receivers, or able to draft on a concrete slab - are tested in various R\&D projects.

Thanks to our close collaboration with major companies in Canada already deploying robotic systems on their construction sites, we tackle some of the main challenges through directed interviews. 

\subsection{Accessibility and acceptance}
Robots are deployed most commonly on a bi-weekly or monthly basis on most of the construction sites, and again, mostly flying robots.

The adoption of robotic systems by workers and the general public varies according to the type of robot used. Drones are becoming more commonly accepted, quite often already seen as common industrial equipment. However, ground robots that operate independently alongside humans are still noticed and trigger skepticism. Workers often criticize ground robots and are afraid of being displaced by them. This significantly affects their acceptance on building sites. The appearance of the ground robots plays a significant role in their acceptance by human workers. More industrial designs are less criticized and better welcomed by the worker community than life-like ones (dog, spider, humanoid, etc). 

\subsection{Deployment Logistics, Operations \& Impact on Schedule}
Regulatory requirements, technological limitations, and location-specific variables all have an impact on logistical issues in the deployment of robotic systems. Drones have become a vital part of the construction industry and are no longer considered as an external or exotic element on construction sites, thus having less deployment complexity. However, the same cannot be said for the ground robots as they are still seen as external elements on a construction site and pose challenges with regulations and acceptance by workers. When considering the autonomy of operations in industry, a notable distinction emerges between ground robots and drones. Ground robots, more often than not, rely on teleoperation, where human operators guide their movements and actions remotely. This is mandated by the regulations when there are human workers on site. Conversely, drones predominantly operate autonomously with waypoint navigation. This autonomy allows drones to execute predefined tasks efficiently, such as surveillance, mapping, or delivery, with minimal human intervention, however, human supervision is mandatory.

During the adoption and developmental phase of robots in construction, the deployment on construction sites may have significant impacts in terms of logistics and impacts on schedules. However, a successful integration of robotics on construction sites can result in increased productivity and safety.

\subsection{Suitability of the construction sites for robotic deployment}
Construction sites differ in their viability for robotic deployment, with building projects being more suitable for drone use due to outside conditions and fewer complicated impediments. Indoor projects involve problems such as increased complexity, greater interaction with people, and security issues, making them less suitable for robotic deployments. However, developments in technology and legal frameworks may provide potential for indoor robotics in the future.

Successful robotic system deployments are dependent on a variety of elements, including technology developments, compliance with safety laws, public trust, and good communication. Advances in battery life and navigation technologies will improve task efficiency and worker protection. Compliance with safety rules guarantees the safe operation of drones and robots, which will promote acceptability among site people. Building public trust via openness and resolving concerns is critical to the widespread deployment of robotic technology in the building sector.

\section{Autonomous mobile robot prototype deployment}
Based on our long-standing collaboration with construction companies already owning robotic systems and testing their deployment on construction sites, we developed a prototype robot, Journeybot, to assess the complexity of these field deployments.

\subsection{The Robot}
\begin{figure}
        \includegraphics[width=\columnwidth]{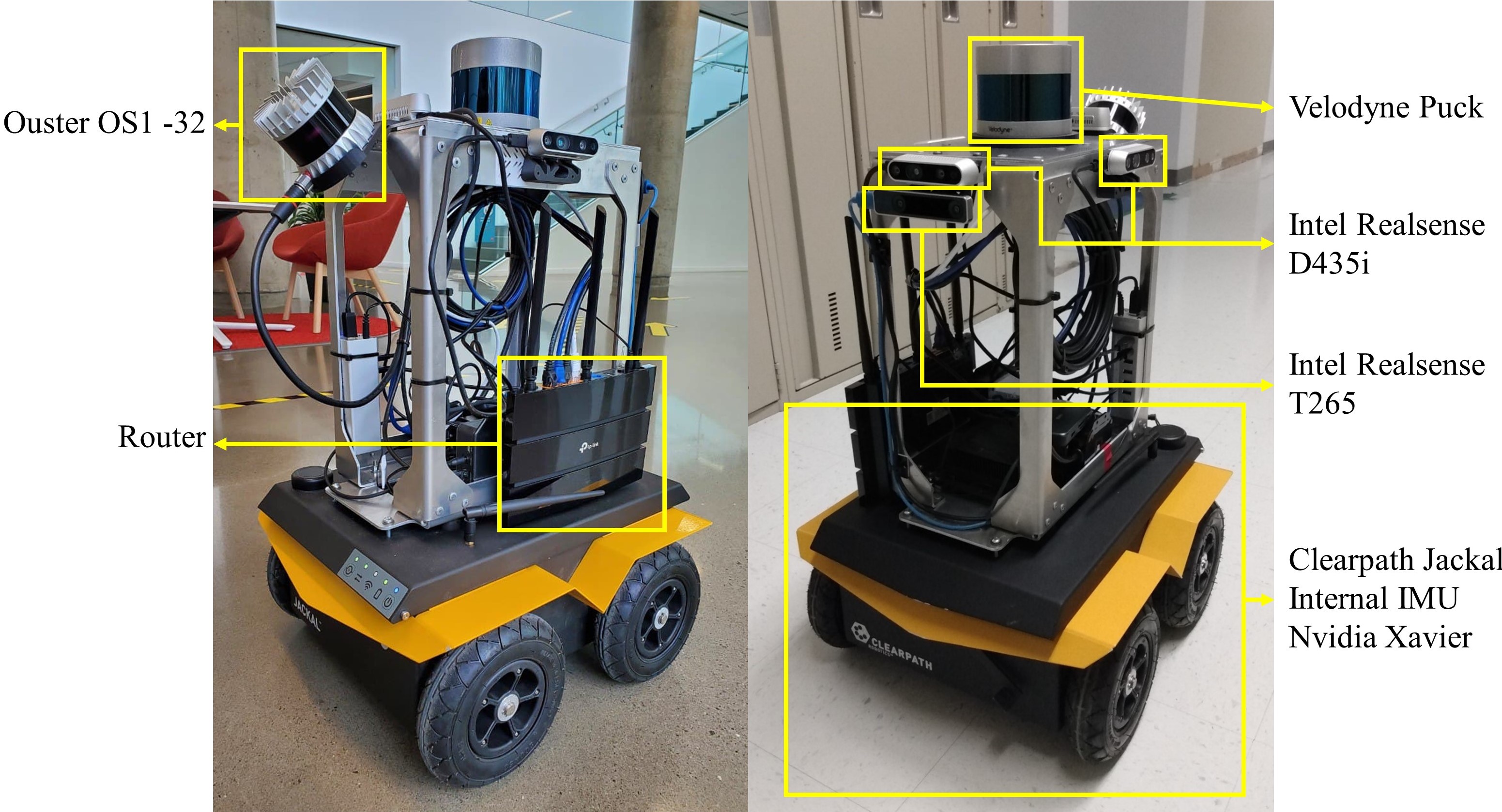}
        \caption{Mobile robot platform equipped with various sensors}
	\label{fig:robotplatform}
\end{figure}

We developed an autonomous mobile robot prototype to perform deployments in construction sites and collect sensor data, named Journeybot. Our robot is built from a four-wheeled unmanned ground vehicle, the Clearpath Jackal, equipped with a hybrid vision/laser scan sensing system as shown in Fig.~\ref{fig:robotplatform}. This robotic system is capable of collecting data while the robot is moving and can navigate semi-autonomously, with a remote operator sending high-level commands. 

The sensing suite of the Journeybot is designed to facilitate comprehensive data collection for digital twin applications. Strategically positioned LiDARs (2) and depth cameras (5) grant a panoramic coverage of the robot’s surroundings for robust environmental perception.

The robot’s pose in the map is obtained through the use of a ROS implementation of the Adaptive Monte Carlo localization algorithm from \cite{thrun2002probabilistic}. Before deploying the robot, wall geometry information is extracted from the Building Information Model (BIM) to generate an occupancy grid of the building. During the robot navigation, this map, the odometry, and the laser scan data from the horizontally mounted Velodyne LiDAR are fed to the localization algorithm, which then estimates the robot’s current pose in that map. When a desired pose command is sent by the remote operator to the robot, a navigation algorithm calculates the shortest path to that pose that avoids all obstacles known from the occupancy grid. This path is constantly updated as the robot moves and new obstacles are detected. More details are given in \cite{karimi2021semantic}.

\subsection{The Fields}

\begin{figure*}
     \centering
     \hfill
     \begin{subfigure}[b]{.55\columnwidth}
         \centering
         \includegraphics[width=\columnwidth]{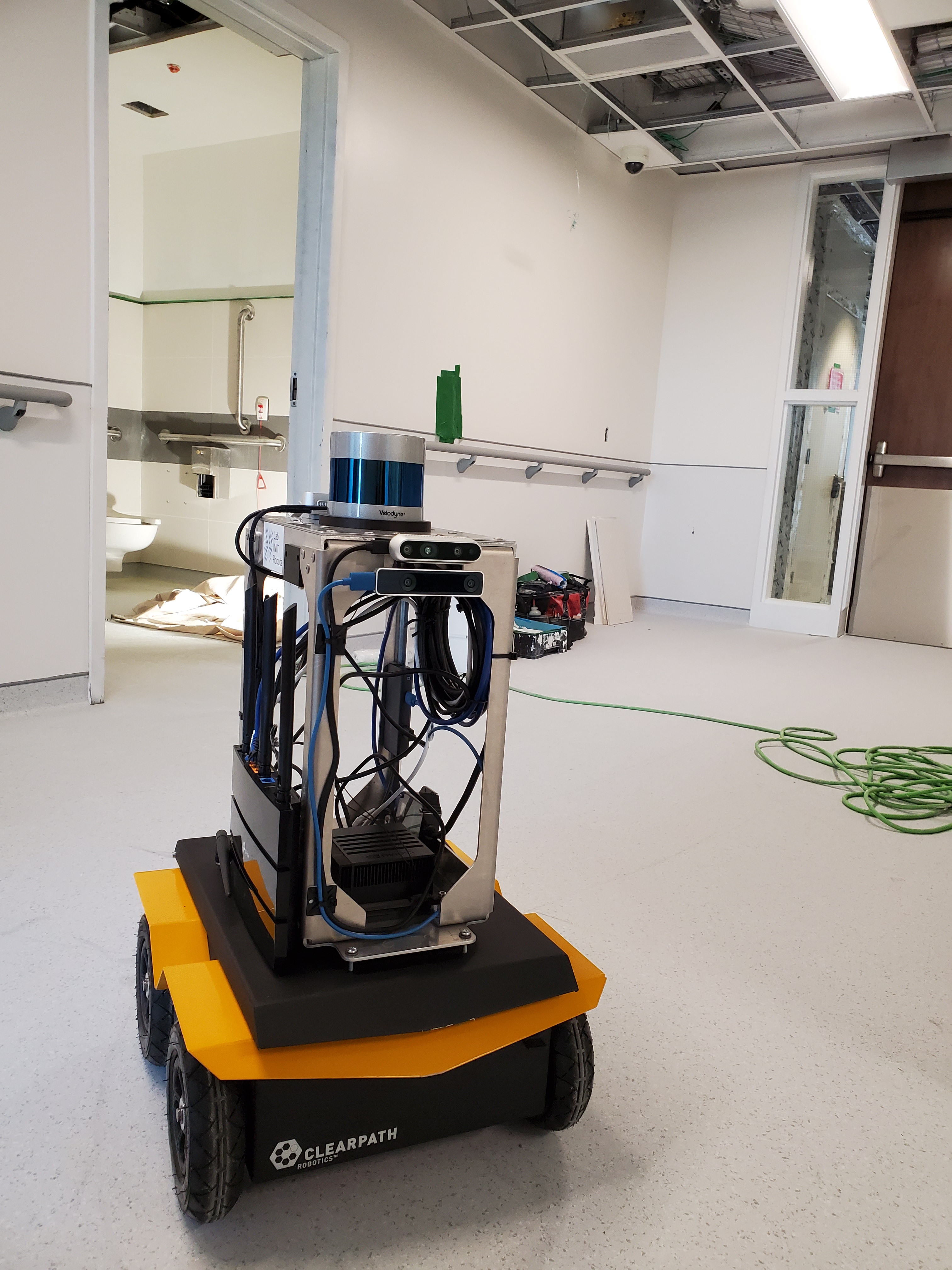}
         \caption{}
         \label{fig:office_floor}
     \end{subfigure}
     \hfill
     \begin{subfigure}[b]{.55\columnwidth}
         \centering
         \includegraphics[width=\columnwidth]{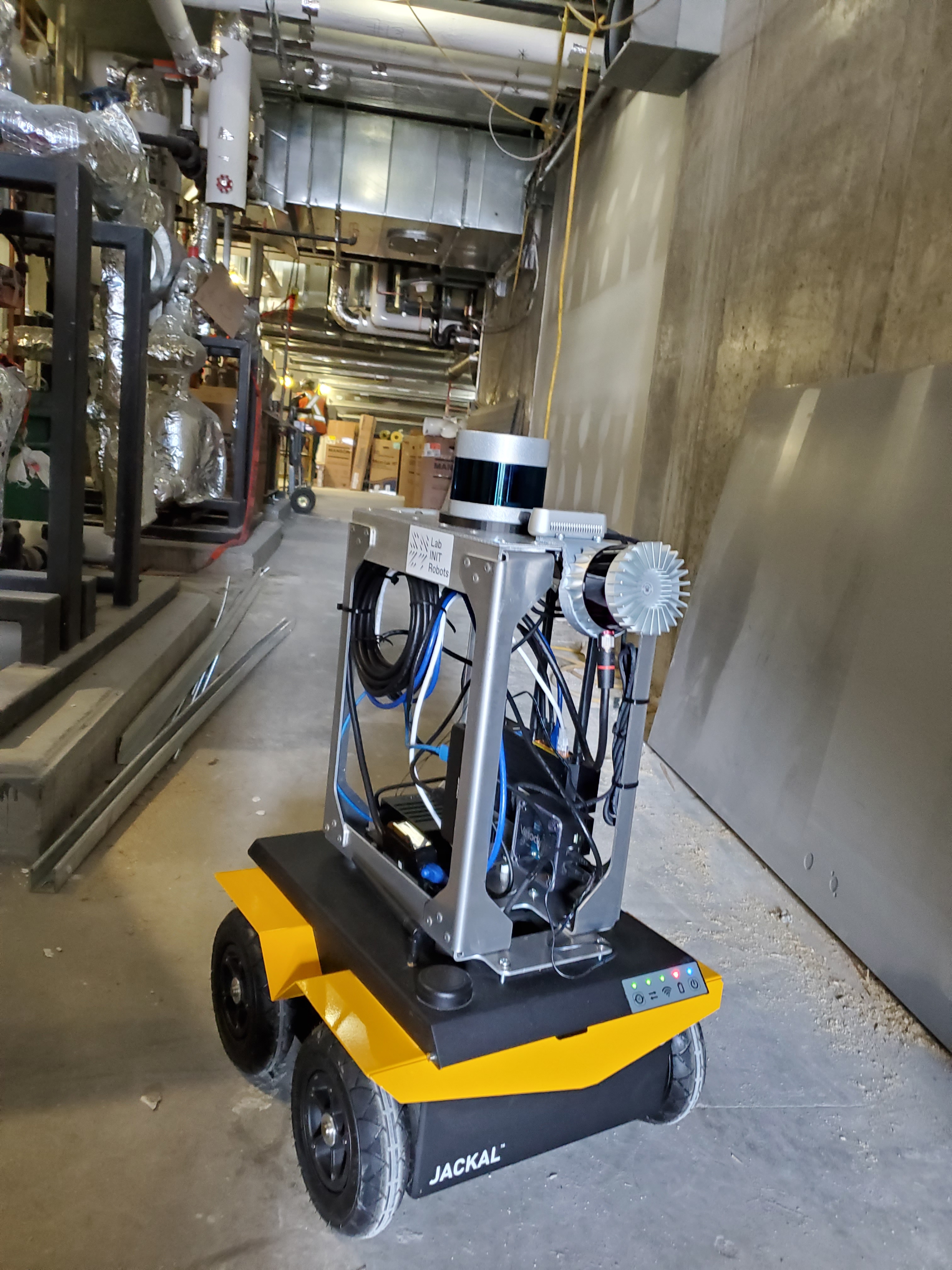}
         \caption{}
         \label{fig:mechanical_floor}
     \end{subfigure}
     \hfill
     \begin{subfigure}[b]{.55\columnwidth}
         \centering
         \includegraphics[width=\columnwidth]{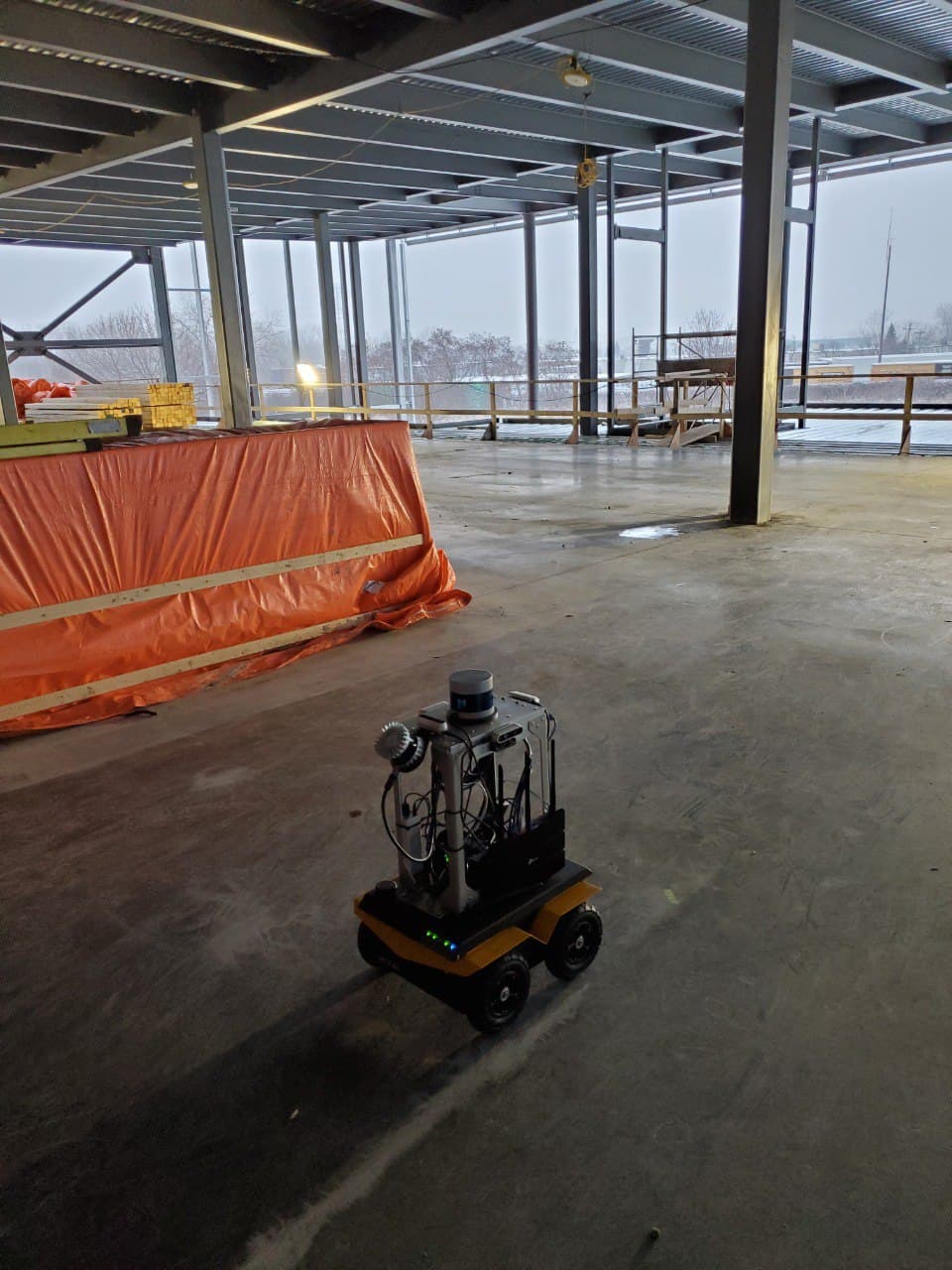}
         \caption{}
         \label{fig:early_stage}
     \end{subfigure}
     \hfill
     \caption{Journeybot in different construction sites: (a) Hospital Building, Office Floor; (b) Hospital Building, Mechanical floor; (c) Second Building, Early Construction Stage}
     \label{fig:construction_sites}
\end{figure*}

Pomerleau Construction based in Montreal, Canada, opened three construction sites to our team, shown in Fig. \ref{fig:construction_sites}. Each deployment presented unique environmental challenges:

\begin{itemize}
    \item Hospital Building, Office Floor: In the later stages of construction, this floor consisted of long, narrow corridors almost clear of construction equipment and materials, although numerous workers were moving around, some occasionally crossing paths with the robot.
    
    \item Hospital Building, Mechanical Floor: In stark contrast to the office floor, this deployment unfolded on a floor dedicated to heating and ventilation equipment. The space was filled by large boilers, electrical generators, water pumps, and piping systems. Despite the absence of construction workers, narrow passages between equipment were challenging.
    
    \item Commercial Building, Early Construction Stage: This early-stage construction site was littered with construction materials and equipment. We deployed the robot on the second floor, where worker presence was minimal, in a vast yet cluttered navigation space. Piles of construction material and assorted obstacles demanded agile maneuvering of the robot.
    
    \item Industrial Building, Middle stage: This factory plant was in the middle stages of construction, with partially installed production lines and scattered construction materials. The larger spaces offered ample room for the robot to navigate easily.
\end{itemize}

Each of these deployments was preceded by a careful analysis by the team of the potential challenges on site.

\subsection{The Logistics}

\subsubsection{Preparation}

Several preparation steps preceding each deployment were instrumental in deploying safely our robotic platform into the dynamic construction environments:

\begin{itemize}
    \item Deployment plan validation: On the days preceding each deployment, we conducted comprehensive testing of the robot's navigation and data collection capabilities within the controlled environment of our university laboratory. We tailored each test from the field deployment layout - trying to reproduce similar trajectories and environments, as much as possible. This involved deploying the robot in several different locations on the campus. Subsequently, we ensured that both the robot's battery and the battery of the remote controller were fully charged and ready for deployment.

    \item Collision Avoidance System testing: Construction site environments require safe management of both equipment and personnel in the robot vicinity. Our autonomous navigation stack includes an emergency collision avoidance system, halting the robot upon detecting anything too close. Rigorous tests are conducted before each deployment to confirm its efficacy.

    \item Assessment of Terrain Compatibility: Our team carefully studied the construction site topography including slopes, stairs, openings, etc. We conducted extensive simulations to confirm the robot's capability to traverse such terrains. From these simulations, we devised contingency plans to manually navigate the robot over obstacles exceeding its capacity.
\end{itemize}

\thispagestyle{withfooter2}

These preparatory measures not only optimized the functionality of our technology but also underscored our commitment to safety and efficiency in real-world applications. These steps may not provide direct insights into research novelty but they help the research team better assess the reality of the application of their research and they ensure better acceptance of the technology on the field (less prone to failure).

\subsubsection{On site}

Upon arriving at the construction sites, we first performed the following procedure:

\begin{itemize}
    \item Present Journeybot to the various stakeholders (workers, supervisors) on the site for them to get to know the new mobile system that will now share their workspace. This step is critical to make sure they know the risks and how to behave in some specific situations with this AGV to reduce the probability of incidents. We let some employees operate the robot and see the sensors' output on its management interface.

    \item Walk through the construction site to survey and validate the path plan and topology studied. On complex fields, we also operate the robot manually for the first run before launching the autonomous navigation stack. It is essential to confirm the functionality of the data acquisition system and that Journeybot is adapted to this environment.
\end{itemize}

During the actual deployment, we sent commands to the robot to explore the environment autonomously and monitor it at all times. One team member walked alongside Journeybot while monitoring its sensors and navigation software output from a laptop that was being carried by one of the engineers. Another team member held the remote controller to take over the robot control if necessary. We would also constantly monitor the environment itself: look out for obstacles and construction workers moving through the environment.

\section{CONCLUSIONS}

Ensuring human safety in shared workplaces with robotic systems presents significant challenges. This paper has explored the safety considerations and productivity impacts of deploying autonomous mobile robots on construction sites. Through our collaborations with leading construction companies in Canada, we have gained valuable industry insights and conducted deployments in real-world scenarios. This experience has enabled us to identify particularly effective strategies. These strategies include performing a thorough site assessment prior to deployment to anticipate potential challenges, testing the robotic system in simulations and environments that closely mimic the actual construction site (BIM-powered), and maintaining checklists to ensure consistency and prevent errors after each deployment. Additionally, maintaining a strong connection with workers to ensure their acceptance of the technology and focusing on functional aesthetics rather than anthropomorphic features have proven essential. Looking ahead, these protocols could be adapted for use with other robotic platforms and applied in a broader range of environments.

\addtolength{\textheight}{-12cm}


\section*{ACKNOWLEDGMENT}

The author thanks CANAM and Pomerleau Construction employees for their precious insights and their generosity in sharing their experience with field deployment of mobile robotic systems.


\bibliographystyle{IEEEtran}
\bibliography{references}

\end{document}